\documentclass[conference]{IEEEtran}
\IEEEoverridecommandlockouts
\usepackage{amsmath,amssymb,amsfonts}
\usepackage{algorithm}
\usepackage{algpseudocode}
\usepackage{graphicx}
\usepackage{textcomp}
\usepackage{xcolor}
\def\BibTeX{{\rm B\kern-.05em{\sc i\kern-.025em b}\kern-.08em
    T\kern-.1667em\lower.7ex\hbox{E}\kern-.125emX}}
\begin{document}

\title{Efficient Fully Distributed Federated Learning with Adaptive Local Links
\thanks{Funded in part by the Project IRIDA under the Cypriot Grant RIF-INFRASTRUCTURES/1216/0017 and in part by the University of Patras.}
}

\author{\IEEEauthorblockN{Evangelos Georgatos}
\IEEEauthorblockA{Comp. Eng. and Informatics Dept. \\
\textit{University of Patras}\\
Patras, Greece \\
egeorgatos@ceid.upatras.gr}
\and
\IEEEauthorblockN{Christos Mavrokefalidis}
\IEEEauthorblockA{Comp. Eng. and Informatics Dept. \\
\textit{University of Patras}\\
Patras, Greece \\
maurokef@ceid.upatras.gr}
\and
\IEEEauthorblockN{Kostas Berberidis}
\IEEEauthorblockA{Comp. Eng. and Informatics Dept. \\
\textit{University of Patras}\\
Patras, Greece \\
berberid@ceid.upatras.gr}
}

\maketitle

\begin{abstract}
Nowadays, data-driven, machine and deep learning approaches have provided unprecedented performance in various
complex tasks, including image classification and object detection, and in a variety of application areas, like autonomous vehicles,
medical imaging and wireless communications. Traditionally, such approaches have been deployed, along with the involved datasets, on
standalone devices. Recently, a shift has been observed towards the so-called Edge Machine Learning, in which centralized  architectures are adopted
that allow multiple devices with local computational and storage resources to collaborate with the assistance of a centralized server. The well-known federated learning
approach is able to utilize such architectures by allowing the exchange of only parameters with the server, while keeping the datasets private to each contributing device.
In this work, we propose a fully distributed, diffusion-based learning algorithm that does not require a central server and propose an adaptive combination rule for the cooperation of the devices. By adopting a classification task on the MNIST dataset, the efficacy of the proposed algorithm over corresponding counterparts is demonstrated via the reduction of the number of collaboration rounds required to achieve an acceptable accuracy level in non-IID dataset scenarios.
\end{abstract}

\begin{IEEEkeywords}
Server-less Federated Learning, Distributed Learning, Diffusion, Adaptive Weights
\end{IEEEkeywords}

\section{Introduction}

Fueled by high data and computational power availability, Machine Learning (ML) has a transformative impact on our lives by offering high accuracy tools for tasks like 
natural language processing, object recognition, medical diagnosis and more \cite{Hatcher2018, zhao2019object, Suzuki2017, Debah}. The ML tasks are traditionally 
implemented in a centralized way either by standalone platforms or via cloud-based architectures, where all the data reside and the computations are performed. However, on 
the one hand, the advent of a new breed of intelligent devices with limited resources \cite{Tang2017} and, on the other hand, time-sensitive applications with high 
reliability requirements, including AR/VR, self-driving vehicles \cite{Debah}, etc., gave rise to the so-called edge ML paradigm \cite{Debah}. The aim of Edge ML is that ML tasks are performed closer to the end-user where the data are truly available. 

Moving towards this paradigm, edge devices (e.g., smart phones, vehicles, base stations) have access to a (possibly) limited in size dataset and they are interested in
learning a particular model to be used for inference in classification or regression tasks. The devices form a network and collaboratively learn
the intended model by exchanging relevant information. The well-known Federated Learning (FL) \cite{GoogleFL} is one of the first approaches to consider edge ML by relying on a, so-called, parameter server that exchanges model parameters with the edge devices but not their data for privacy reasons. 

Currently, there are considerable research efforts that aim to address many of the challenges identified in the frame of FL \cite{FLadvances}. Notably, one major challenge in FL is the existence of non Independently and Identically Distributed (non-IID) datasets among the devices, namely, the datasets of particular devices are not representative of the desired task, e.g., in a classification task, data for certain classes are missing. As a remedy to this problem, it has been proposed in literature that FL may be improved 
by exchanging, apart from parameters, gradient information as well \cite{GoogleFL,FLnonIIDdata,CHINESE}. Additionally, the central server can be a single point of failure, prone to adversarial attacks and a bottleneck when a large number of devices are involved. There have been some works towards avoiding such a server altogether by adopting fully distributed approaches like \cite{FLarch,ITALIANS,cross-grad} and references therein.
 
In this paper, we propose a fully distributed, data-driven learning algorithm and a novel adaptive combination rule. In particular, the highlights of the paper are as follows. Capitalizing on the distributed (adaptive) signal processing literature \cite{sayed2014adaptation,plata2015distributed}, a distributed learning algorithm of the diffusion-type is introduced where combinations of information among neighboring devices are performed via constant weights. Additionally, a novel adaptive combination rule is devised for the proposed diffusion-based algorithm based on gradient information leading to improved learning rate convergence especially for the challenging non-IID dataset case, with low and constant communication overhead that does not depend on the formed neighborhood sizes as in \cite{ITALIANS,cross-grad}. Finally, the efficacy of  the proposed distributed algorithm with constant and adaptive combinations is demonstrated via a classification task in the MNIST dataset \cite{MNIST} in terms of non-IID datasets and varying number of participating devices.

In the rest of the paper, Sec.~II
presents the system model and formulates the problem under consideration. In Sec.~III, the proposed fully-distributed algorithm is described along with the adaptive combination rule and a discussion on the communication complexity is provided. Finally, in Sec.~IV, the performance  of the proposed algorithm is assessed via extensive simulations and Sec.~V concludes the paper.

\section{System Model and Problem Formulation}

In this section, first, the underlying system model for the network of the cooperating devices (or agents) will be presented and, then, the optimization problem that is to be solved by the network, will be formulated.

\subsection{System Model}

\begin{figure}[t]
  \centering
  \includegraphics[width=\columnwidth,keepaspectratio]{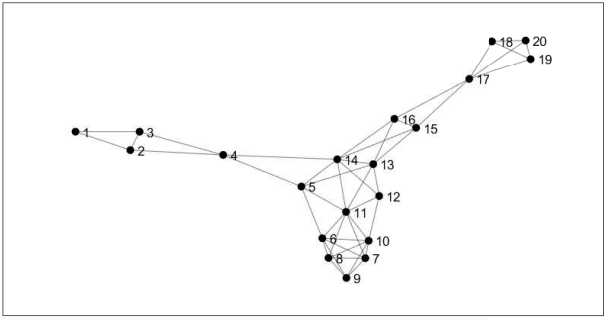}
  \vspace{-2cm}
  \caption{Network of $N$=20 agents.}
  \label{fig:example_net}
\end{figure}

In this work, a strongly connected network of $N$ agents is assumed. The network is modeled as a graph $G(\mathcal{N},\mathcal{E})$, where $\mathcal{N}$ is the set of nodes that represent the agents and $\mathcal{E}$ is the set of edges that connect pairs of agents being able to exchange information (as an example see Fig.~\ref{fig:example_net}). The neighborhood of
agent $k\in \mathcal{N}$ is depicted by the set $\mathcal{N}_k$ and consists of all agents $j$ for which the edge $(k,j)\in \mathcal{E}$, including agent $k$.  Similarly, $\mathcal{N}_{\bar{k}}$ is the neighborhood of $k$, excluding this time the agent $k$. 
 
\subsection{Problem Formulation}

The aim of the network is to solve in a collaborative way a stochastic optimization problem for a task of common interest by allowing each agent to interact with its neighborhood. In such problems, agent $k$ aims at minimizing a stochastic risk function of the form
\begin{equation}
    F_k(\mathbf{w}) = \mathbb{E}\{l(\mathbf{w};\mathbf{x}_k)\}, 
    \label{eq:stochastic_risk}
\end{equation}
where $l(\mathbf{w};\mathbf{x}_k)$ is the loss function, $\mathbf{w}\in \mathbb{R}^M$ is a vector consisting of the model parameters, relevant to the task at hand (e.g., in a classification problem, these parameters could be the weights of a neural network), $\mathbf{x}_k$ refers to the stochastic data and $\mathbb{E}\{\cdot\}$ refers to statistical expectation over the distribution of the data. Then, the network aims at minimizing the overall problem  
\begin{equation}
 \min_{\mathbf{w}} F(\mathbf{w}),\textrm{\,where\,} F(\mathbf{w}) = \sum_{k=1}^Np_kF_k(\mathbf{w}),
 \label{eq:min_prob}
\end{equation}
where $p_k\geq 0$ sum up to one. 

Under the assumption that the statistical distribution of the data is unknown, agent $k$ utilizes dataset $\mathcal{D}_k$ with $|\mathcal{D}_k|=D_k$
elements and substitutes the stochastic risk function in \eqref{eq:stochastic_risk} with the corresponding empirical risk function that approximates the statistical expectation with an arithmetic mean of the loss function $l(\cdot)$ over the $D_k$ elements of the dataset. In this case, $p_k=D_k/D$ (where $D=\sum_kD_k$) which is simplified to $p_k=1/N$ when the data are uniformly distributed among the agents, i.e., $D_k=D_j$, $\forall, i,j$. Finally, it is noted that under the IID and non-IID scenarios, the data elements of all $\mathcal{D}_k$'s are produced according to same and different underlying statistical distributions \cite{FLnonIIDdata,FLadvances}, respectively. In the latter case, considering a classification task as an example, this may lead to datasets that do not capture any data for certain classes \cite{ITALIANS}.

\section{Fully Distributed Federated Learning}

In this section, first, fully distributed federated learning is addressed by proposing a distributed diffusion-based algorithm with constant combination weights. Then, an adaptive combination rule is proposed in which each agent utilizes its neighbors' gradient information to determine the combination weights at each communication round. It is noted that this rule, although designed here for the proposed diffusion algorithm, it can also be applied to other distributed algorithms like the one (of the consensus type) proposed in \cite{ITALIANS}. Finally, the communication cost of the proposed algorithm with constant and adaptive weights is analysed
in terms of the required information exchanges and compared with other relevant variants.

\subsection{Distributed Diffusion-based Federated Learning}

Driven by previous works on distributed parameter estimation \cite{DifAdt}, the proposed algorithm adopts the diffusion approach for enabling the agents to collaboratively solve \eqref{eq:min_prob} and builds upon the so-called Adapt-then-Combine (ATC) strategy. This strategy has been shown to provide enhanced performance and stability guarantees when compared to the Combine-then-Adapt alternative \cite{sayed2014adaptation}, \cite{DifAdt}. 

Before describing the proposed algorithm in detail, let us first present the two main steps involved generally in the ATC strategy, which are the following.
\begin{equation}
 \boldsymbol{\psi}_{k,t} = \mathbf{w}_{k,t-1} -\mu_{k} \hat{\nabla} F_{k}(\mathbf{w}_{k,t-1})
 \label{eq:ATCAdapt}
 \end{equation}
 \begin{equation}
       \mathbf{w}_{k,t} = \sum_{l \in \mathcal{N}_k} a_{lk}\boldsymbol{\psi}_{l,t}
       \label{eq:ATCCombine}
 \end{equation}
In the first step (see \eqref{eq:ATCAdapt}), agent $k$ adapts its local parameter vector $\mathbf{w}_{k,t-1}$ using a Stochastic Gradient Descent (SGD) update and produces an intermediate 
vector $\boldsymbol{\psi}_{k,t}$. To achieve this, an approximation $\hat{\nabla} F_k$ of the true gradient $\nabla F_k$ is employed with a step size $\mu_k$. Then, in the second step (see \eqref{eq:ATCCombine}), after receiving  the corresponding intermediate parameter vectors from its neighbors, agent $k$ employs a convex combination
of all intermediate vectors using the weights $a_{lk}$, $l\in \mathcal{N}_k$, which are non-negative and add up to one, and, thus, the new parameter vector  $\mathbf{w}_{k,t}$ is created. Note that in the context of FL, the parameter vector is actually the target model (e.g., a neural network) for the desired task (e.g., classification).

%At step (2b) each agent receives the intermediate estimations of his neighbors and combines them. The $a_lk$ are convex combination weights satisfying
%\[\sum_{l \in N_k}a_{lk} = 1.\]

The proposed distributed diffusion-based FL algorithm assumes that each agent $k$ employs the ATC strategy $T$ times (or communication rounds) during an epoch (which are $E$ in total). In communication round $t$, agent $k$ adapts its local model using SGD updates for a number of mini-batches of size $b$ (selected randomly from $\mathcal{D}_k$). After that, agent $k$ sends its intermediate model  $\boldsymbol{\psi}_{k,t}$ (and $\boldsymbol{\delta}_{k,t}=\boldsymbol{\psi}_{k,t}-\mathbf{w}_{k,t-1}$ when the adaptive combination rule is employed) to its neighbors while receiving the corresponding quantities from the latter. Finally, agent $k$ performs the combination step to acquire the updated model $\mathbf{w}_{k,t}$. Note that $\boldsymbol{\delta}_{l,t}$'s, $l\in\mathcal{N}_k$, provide information about the local gradients of the neighborhood of agent $k$ and they are used for determining the weights for the adaptive combination rule, as will be explained in the next section. The whole process for fully-distributed diffusion-based FL is presented in Algorithm \ref{alg:cap}.
 
% The same policy used by federated averaging algorithm \cite{GoogleFL} is used and here for the selection of $a_{lk}$ combination weights. A constant rule is used based on each agent's dataset size.
As a final note, in this work, for the case of constant combination weights, the following rule (adopted commonly in the relevant literature \cite{GoogleFL}, \cite{ITALIANS}) is applied 
   \begin{equation}
          a_{lk} = \frac{D_l}{\sum_{j\in \mathcal{N}_{k}}D_j},
          \label{eq:constant}
   \end{equation}
which depends on the involved agents' dataset sizes. However, other rules for determining constant combination weights can be employed like the ones proposed in \cite{SAYEDBOOKCHAPTER}.

\begin{algorithm}[t]
\caption{\textbf{Distributed Diffusion-based FL}}\label{alg:cap}
\begin{algorithmic}[1]
\Procedure{Diffusion Learning}{$N_{k},T,E,\{\mu_l\},B$}
\State Initialize: $\mathbf{w}_{k,0}, t = 0$
\For{each epoch $e = 1,2,...,E$}
 \For{each communication round $n = 1,2,...,T$}
     \State $t\gets t+1$
     \State   $\boldsymbol{\psi}_{k,t} \gets Adapt(\mathbf{w}_{k,t-1})$\Comment{Adapt Step}
     \If{ConstantWeights}
     \State Send($\boldsymbol{\psi}_{k,t}$)
     \State Receive$\left(\{\boldsymbol{\psi}_{l,t}\}_{l \in \mathcal{N}_{\bar{k}}}\right)$
      \State Use \eqref{eq:constant} to set $a_{lk}$'s 
     \ElsIf{AdaptiveWeights}
     \State $\boldsymbol{\delta}_{k,t} = \boldsymbol{\psi}_{k,t} - \mathbf{w}_{k,t-1}$
     \State Send($\boldsymbol{\delta}_{k,t},\boldsymbol{\psi}_{k,t}$)
     \State Receive$\left(\{\boldsymbol{\delta}_{l,t}\}_{l \in \mathcal{N}_{\bar{k}}},\{\boldsymbol{\psi}_{l,t}\}_{l \in \mathcal{N}_{\bar{k}}}\right)$
      \State Use \eqref{eq:adaptiveweights} to set $a_{lk}$'s 
     \EndIf
     \State $\mathbf{w}_{k,t} \gets \sum_{l \in \mathcal{N}_k} a_{lk}\boldsymbol{\psi}_{l,t}$\Comment{Combine Step}
      \EndFor
      \EndFor
\EndProcedure
\Procedure{Adapt}{$\mathbf{w}_{k,t-1}$}
\State $ \mathcal{B} \gets \text{mini-batches of size b}$
\State $\boldsymbol{\psi}_{k,t} \gets SGD(\mu_k,\mathcal{B},\mathbf{w}_{k,t-1})$
\EndProcedure
\end{algorithmic}
\end{algorithm}

 \subsection{Adaptive Gradient-based Combination Rule}
 
 The proposed distributed algorithm, as other learning algorithms in the machine learning literature, relies on SGD updates that employ an approximate gradient that is computed over the available dataset. In order the approximate gradient to be an unbiased estimate of the true gradient, the SGD algorithm assumes (and requires) an IID sampling of the data comprising the involved dataset \cite{CHINESE}. 
 
 However, this assumption might be problematic in the FL or fully distributed learning cases as such data sampling could be difficult to achieve. The main reason of non-IID datasets in distributed settings is the fact that they are probably captured by agents with different geographical and temporal particularities \cite{FLadvances}. The existence of non-IID data has undesirable effects during the training phase (when a model is learned by solving, e.g., minimization problems like \eqref{eq:min_prob}), leading to reduced convergence speed or even diverging from the desired model \cite{CHINESE}. This behavior is a consequence of the
 close relation between each agent's local model $\mathbf{w}_k$ and the statistical distribution of its data. A large number of local updates will lead towards a local minimum of the loss function $F_k(\mathbf{w}_k)$  that will be inconsistent with the desired global solution. 
 This inconsistency is accumulated during training leading to an increased number of communication rounds before convergence is achieved.
 
 Taking this behavior into account, the convergence speed of the learning procedure can be improved via more sophisticated combination rules for determining the weights $a_{lk}$. To this end, an adaptive rule is proposed building upon the approach devised in \cite{CHINESE} for the classical FL case where a parameter server exists. The main idea for this rule is that gradient information is taken into account at each communication round in order to assess each agents' ``divergence'' from the optimal minimization path. 
 In the frame of FL, this is achieved by the parameter server via comparing its approximate global gradient with local approximate gradients sent by the agents and weighing the contribution of each agent accordingly. However, this is not possible in fully distributed learning settings. Instead, during a communication round, agent $k$ approximates the global gradient in terms of its neighborhood by aggregating the involved gradients as follows. 
\begin{equation}
         %\nabla F_k(\mathbf{w}_{k,t}) = \sum_{l \in \mathcal{N}_{k} } \frac{|D_l|}{\sum_{l \in N_{k}}|D_l|} \nabla F_{l}(w_{t,l})  
         \hat{\nabla} F^g_k(\mathbf{w}_{k,t}) = \sum_{l \in \mathcal{N}_{k} } a_{lk} \hat{\nabla} F_{l}(\mathbf{w}_{l,t-1})=-\sum_{l \in \mathcal{N}_{k} } a_{lk} \frac{\boldsymbol{\delta}_{l,t}}{\mu_l}.
         \label{eq:totalgrad}
     \end{equation}
 The $a_{lk}$'s are defined as in \eqref{eq:constant}, while it is reminded that the $\boldsymbol{\delta}_{l,t}$'s, $l\in \mathcal{N}_k$, have been sent to agent $k$. 
 %and that $\hat{\nabla} F_{l}(\mathbf{w}_{l,t}) =  - \frac{\boldsymbol{\delta}_{l,t}}{\mu_l}$.

The comparison of the approximate global gradient in \eqref{eq:totalgrad} with the local gradient of agent $l\in \mathcal{N}_k$,  is performed in terms of the angle between the involved vectors. This is determined using the well-known dot-product equation as follows.
\begin{equation}
         %\theta_{k,l}(t) = \arccos{\frac{<\hat{\nabla}^\textrm{T} F^g_k(w_{t,k}),\nabla F_l(w_{l,t})>}{\|\nabla F_k(W_t)\| \|\nabla F_l(W_{l,t})\|}}
         \theta_{k,l}(t) = \arccos\left(\frac{-\boldsymbol{\delta}^\textrm{T}_{l,t}\hat{\nabla} F^g_k(\mathbf{w}_{k,t})}{\|\boldsymbol{\delta}_{l,t}\|\|\hat{\nabla} F^g_k(\mathbf{W}_t)\|}\right),
         \label{eq:cosim}
     \end{equation}
where $(\cdot)^\textrm{T}$ is the transpose operator and $\|\cdot\|$ is the Euclidean norm.

For the eventual determination of the weights, the remaining steps are the same as the ones in \cite{CHINESE} and they are presented in the following for completeness. Thus, first, a smoothed version of $\theta_{k,l}(t)$ is used to deal with its random nature.
\begin{equation}
         \tilde{\theta}_{k,l}(t)=\frac{t-1}{t}\tilde{\theta}_{k,l}(t-1) + \frac{1}{t}\theta_{k,l}(t),\, t\geq 1 
% \begin{cases}
% \theta_{k,l}(t),  \:   t = 1\\
% \frac{t-1}{t}\Tilde{\theta_{k,l}}(t-1) + \frac{1}{t}\theta_{k,l}(t) \: t > 1\\
% \end{cases}
\label{eq:smoothedangle}
     \end{equation}
Then, the adaptive combination rule for determining the time-varying 
weights $a_{lk}(t)$ is given as
% \begin{equation}
%       a_{lk}(t) = \begin{cases}
%       \frac{e^{f(\tilde{\theta}_{k,l}(t))}}{\sum^{N_k}_{j=1}e^{f(\tilde{\theta}_{k,j}(t))}}, D_m = D_n,\: \forall m,n \in \mathcal{N}_k\\
%       \frac{D_l e^{f(\tilde{\theta}_{k,l}(t))}}{\sum^{N_k}_{j=1}D_{j} e^{f(\tilde{\theta}_{k,j}(t))}}, \: D_m \ne D_n \: \exists m,n \in \mathcal{N}_k 
%       \end{cases}
%       \label{eq:adaptiveweights}
% \end{equation}
\begin{equation}
      a_{lk}(t) = \frac{D_l e^{f(\tilde{\theta}_{k,l}(t))}}{\sum^{N_k}_{j=1}D_{j} e^{f(\tilde{\theta}_{k,j}(t))}},
      \label{eq:adaptiveweights}
\end{equation}
where $N_k=|\mathcal{N}_k|$, $f(x)=a(1 - e^{-e^{-a(x-1)}})$ is a variant of the Gombertz function and $a$ is a scalar \cite{CHINESE}. Note that \eqref{eq:adaptiveweights} does not
depend on the dataset sizes when the datasets are balanced, i.e., of the same size.

% The adaptive weighting assignment on each node requires 2 stages. First, a non-linear mapping of the smoothed angle is considered based on the Gombertz function[].
% \begin{equation}
%       f(\Tilde{\theta_{k,l}}(t)) = a(1 - e^{-e^{-a(\Tilde{\theta_{k,l}(t)-1})}})
%       \label{eq:gombertz}
% \end{equation}

%The final weight calculation is based on the Softmax function as follows

Finally, it is mentioned that the proposed adaptive combination rule can be also applied to other fully distributed FL algorithms like the one in \cite{ITALIANS} which is of the consensus type. In this case, a minor modification is required as agent $k$ combines information from agents in $\mathcal{N}_{\bar{k}}$, i.e., excluding itself. Therefore, the computations of \eqref{eq:totalgrad}-\eqref{eq:adaptiveweights} can be adjusted appropriately.

\subsection{Communication Cost Analysis}\label{sec:complexity}

The proposed algorithm and the consensus-based one in \cite{ITALIANS} require from agent $k$ to broadcast to its neighborhood the local model (with $M$ parameters) at each communication round when the constant weights are used. On the other hand, when the adaptive weights are employed in both algorithms, agent $k$ broadcast both the local model and the local approximate gradient, thus, $2M$ parameters need to be transmitted. 
In the literature, other relevant methods can be found that utilize gradient information like \cite{ITALIANS} and \cite{cross-grad} and references therein.
To the best of our knowledge, in all cases, the communication overhead is much higher and depends on the neighborhood size of agent $k$, thus, leading to
the transmission of $O(MN_{\bar{k}})$ parameters, where $N_{\bar{k}}=|\mathcal{N}_{\bar{k}}|$. For this reason, they are not considered in the following simulations.

%The main interest lies on keeping the communication overhead low. For this reason, in section IV only the CFA algorithm \cite{ITALIANS} is considered as it exchanges the same amount of parameters with the proposed diffusion methods. In exact, those methods exchange their local models on every communication round.

\section{Simulations}

In this section, the performance of the proposed diffusion-based FL algorithm employing the constant and adaptive combination rules, is studied. In the simulations, the consensus-based federated averaging algorithm of \cite{ITALIANS} (properly modified so as to incorporate the same weighting rules) is also presented. Furthermore, for benchmarking purposes, the performance of centralized and individual training is also depicted, namely, when all data are gathered in a central server which performs the training and when each agent operates individually with no cooperation, respectively. For the evaluation of the performance, the task
of classification on the well-known MNIST dataset is considered (consisting of 60K training and 10K testing images of handwritten digits), in which the classifier is a fully connected neural network with two hidden layers
as in \cite{FLarch} and cross-entropy loss is used. For all implementations, MATLAB$^\copyright$ 2021b and the deep learning toolbox have been used.

The performance is investigated via Monte-Carlo simulations adopting a challenging non-IID scenario and the mean testing accuracy over the network of agents is depicted versus the epochs. At each instance (out of $10$ in total), the status of every agent (i.e., whether it has an IID dataset or not) and
which classes compose the non-IID datasets, are selected in random. Subsequently, for each IID agent, the dataset is randomly selected from the whole training set of MNIST images, while, for the non-IID agents, the corresponding datasets are randomly selected by considering only the selected classes per agent. At the end of each epoch, mean accuracy is measured using the MNIST testing images. For the simulations, two networks of size $N=4$ (see the simple linear network in \cite{ITALIANS}) and $N=20$ (see \cite{OLFATI-SABER} and Fig.~\ref{fig:example_net}), are considered. Furthermore, $\mu_k=0.01$ and $D_k= 600$, $\forall k$, while $b=10$, $T=6$, $E=30$, and $a = 5$ in \eqref{eq:adaptiveweights} \cite{CHINESE}.

\begin{figure}[h]
  \centering
  \includegraphics[width=\columnwidth,keepaspectratio]{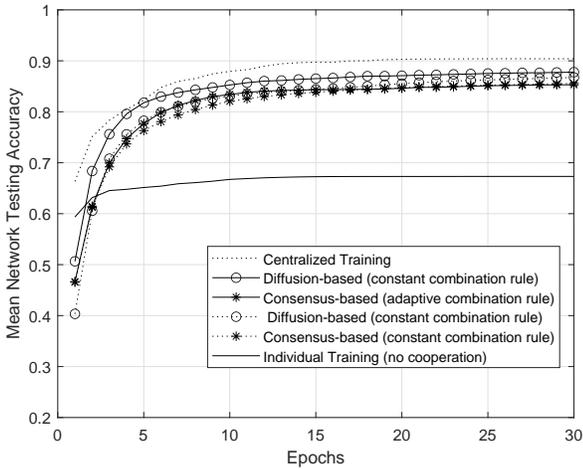}
  \caption{$N=4$ agents with non-IID datasets.}
  \label{fig:4noniid}
\end{figure}

\begin{figure}[h]
  \centering
  \includegraphics[width=\columnwidth,keepaspectratio]{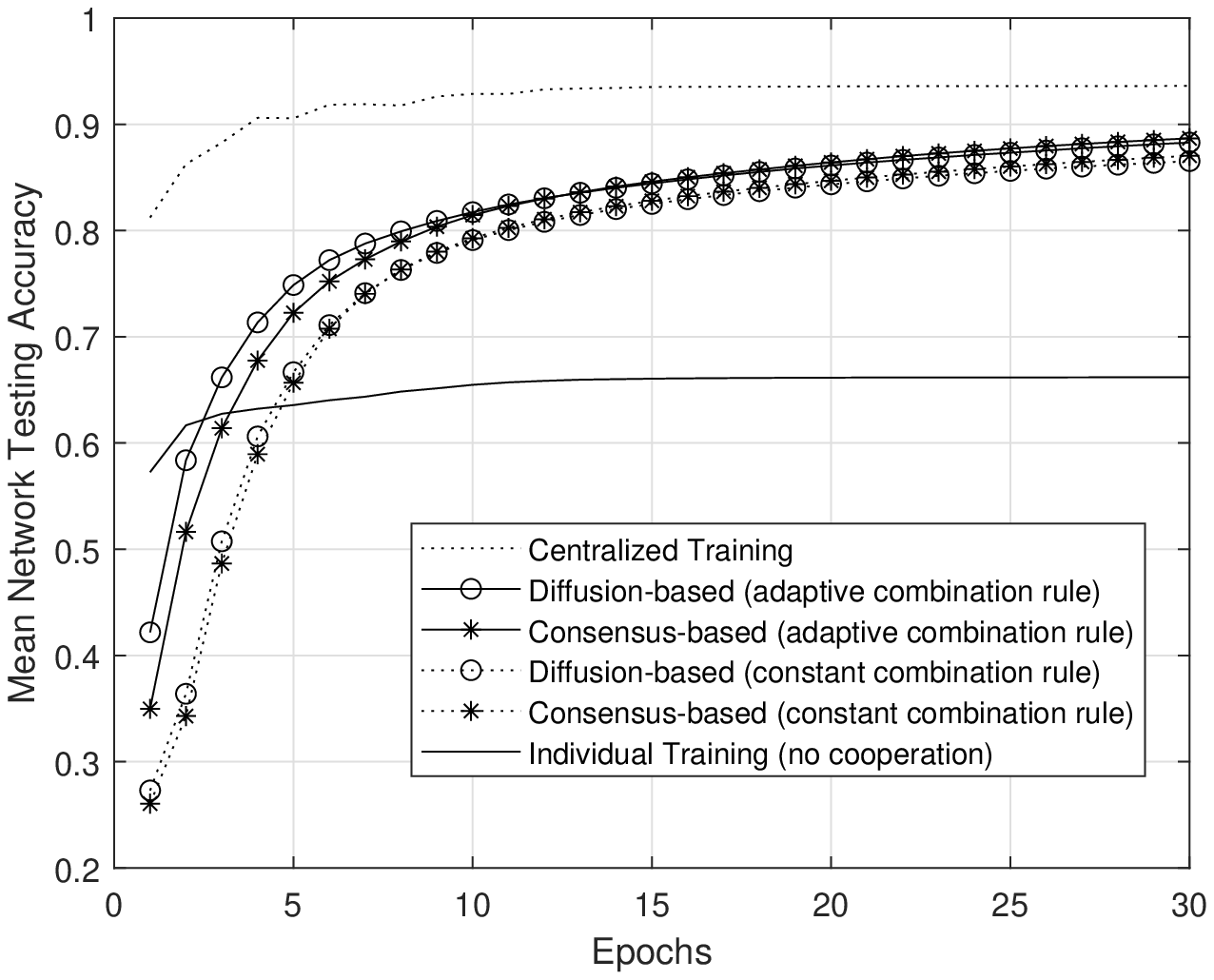}
  \caption{$N=20$ agents with non-IID datasets.}
  \label{fig:20noniid}
\end{figure}

In Fig.~\ref{fig:4noniid}, the results for the network with the $N=4$ agents is depicted. As expected, the distributed algorithms exhibit improved  performance when compared with the no cooperation alternative. Furthermore, the proposed algorithm with adaptive weights demonstrates faster convergence speed versus the other variants, meaning that it reaches a certain accuracy level with less epochs, which leads to less communication rounds (it is reminded that $T=6$ rounds per epoch are employed here). Similarly, in Fig.~\ref{fig:20noniid}, the results for the larger network with $N=20$ agents,
is depicted. Here, the network is composed of richer neighborhoods and more cooperating agents. Thus, better testing accuracy can is achieved. Moreover, in this larger network, the algorithms with the adaptive rules are better than the ones with the constant rule, while the diffusion-based one with adaptive weights demonstrates faster convergence, especially, during the first epochs.  

\begin{figure}[h]
  \centering
  \includegraphics[width=\columnwidth,keepaspectratio]{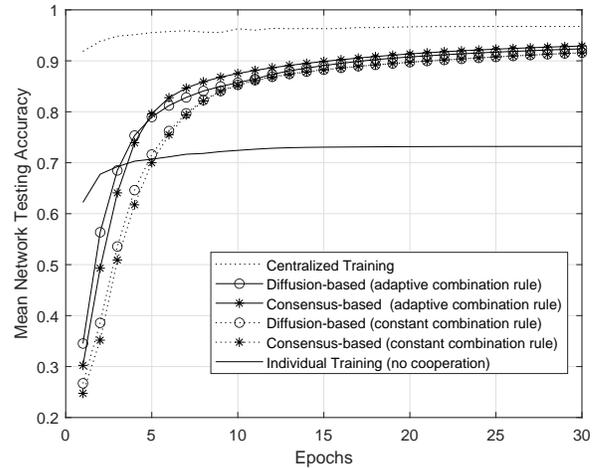}
  \caption{5 central agents (out of $20$) with non-IID datasets.}
  \label{fig:central}
\end{figure}

\begin{figure}[h]
  \centering
  \includegraphics[width=\columnwidth,keepaspectratio]{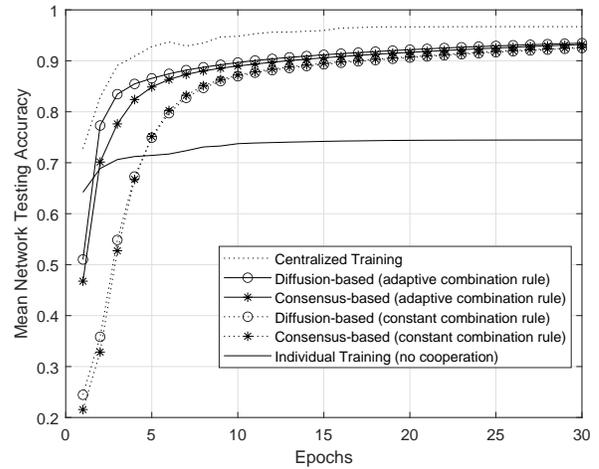}
  \caption{5 edge agents (out of $20$) with non-IID datasets.}
  \label{fig:edge}
\end{figure}

Finally, in Figs.~\ref{fig:central},~\ref{fig:edge}, the performance of the algorithms is demonstrated for two specific distributions of the non-IID agents on the network with $N=20$. In particular, in Fig.~\ref{fig:central}, the five central agents (namely, 5, 11, 12, 13, 14 in Fig.\ref{fig:example_net}) are non-IID and observe at most five (out of ten) classes. Similarly, in Fig.~\ref{fig:edge}, five edge nodes (namely, 1, 20, 9, 8, 19) are now non-IID. As observed, the adaptive rule is crucial on the performance of the distributed algorithms while the behavior of the diffusion-based and consensus-based counterparts is
reversed. This is an interesting remark that was not observed in the previous mean results and deems further investigation in the near future. 

%\section*{Acknowledgment}

%\color{red}To be written\color{black}
\section{Conclusions}

In this work, the problem of fully distributed diffusion-based FL has been studied. Also, a novel adaptive combination rule
has been devised that speeds up the convergence rate of learning, especially under the challenging non-IID dataset scenario.
The performance of the proposed algorithm has been assessed in comparison with other counter-parts of the same communication complexity
in the context of a typical classification task using the well-known MNIST dataset.

%\printbibliography %Prints bibliography

\bibliographystyle{IEEEtran}
\bibliography{references}

\begin{thebibliography}{10}
\providecommand{\url}[1]{#1}
\csname url@samestyle\endcsname
\providecommand{\newblock}{\relax}
\providecommand{\bibinfo}[2]{#2}
\providecommand{\BIBentrySTDinterwordspacing}{\spaceskip=0pt\relax}
\providecommand{\BIBentryALTinterwordstretchfactor}{4}
\providecommand{\BIBentryALTinterwordspacing}{\spaceskip=\fontdimen2\font plus
\BIBentryALTinterwordstretchfactor\fontdimen3\font minus
  \fontdimen4\font\relax}
\providecommand{\BIBforeignlanguage}[2]{{%
\expandafter\ifx\csname l@#1\endcsname\relax
\typeout{** WARNING: IEEEtran.bst: No hyphenation pattern has been}%
\typeout{** loaded for the language `#1'. Using the pattern for}%
\typeout{** the default language instead.}%
\else
\language=\csname l@#1\endcsname
\fi
#2}}
\providecommand{\BIBdecl}{\relax}
\BIBdecl

\bibitem{Hatcher2018}
W.~G. Hatcher and W.~Yu, ``A survey of deep learning: platforms, applications
  and emerging research trends,'' \emph{IEEE Access}, vol.~6, pp.
  24\,411--24\,432, 2018.

\bibitem{zhao2019object}
Z.-Q. Zhao, P.~Zheng, S.-t. Xu, and X.~Wu, ``Object detection with deep
  learning: A review,'' \emph{IEEE transactions on neural networks and learning
  systems}, vol.~30, no.~11, pp. 3212--3232, 2019.

\bibitem{Suzuki2017}
K.~Suzuki, ``Overview of deep learning in medical imaging,'' \emph{Radiological
  physics and technology}, vol.~10, no.~3, pp. 257--273, 2017.

\bibitem{Debah}
J.~Park~et al., ``Wireless network intelligence at the edge,''
  \emph{Proceedings of the IEEE}, vol. 107, no.~11, 2019.

\bibitem{Tang2017}
J.~Tang, D.~Sun, S.~Liu, and J.-L. Gaudiot, ``Enabling deep learning on iot
  devices,'' \emph{Computer}, vol.~50, no.~10, pp. 92--96, 2017.

\bibitem{GoogleFL}
B.~McMahan~et al., ``Communication-efficient learning of deep networks from
  decentralized data,'' in \emph{Artificial intelligence and statistics}.\hskip
  1em plus 0.5em minus 0.4em\relax PMLR, 2017, pp. 1273--1282.

\bibitem{FLadvances}
P.~Kairouz~et al., ``Advances and open problems in federated learning,''
  \emph{Foundations and Trends{\textregistered} in Machine Learning}, vol.~14,
  no. 1--2, 2021.

\bibitem{FLnonIIDdata}
Y.~Zhao, M.~Li, L.~Lai, N.~Suda, D.~Civin, and V.~Chandra, ``Federated learning
  with non-iid data,'' \emph{arXiv preprint arXiv:1806.00582}, 2018.

\bibitem{CHINESE}
H.~Wu and P.~Wang, ``Fast-convergent federated learning with adaptive
  weighting,'' \emph{IEEE Transactions on Cognitive Communications and
  Networking}, vol.~7, no.~4, pp. 1078--1088, 2021.

\bibitem{FLarch}
Y.~Qu~et al., ``Serverless federated learning for uav networks: Architecture,
  challenges, and opportunities,'' \emph{arXiv e-prints}, 2021.

\bibitem{ITALIANS}
S.~Savazzi, M.~Nicoli, and V.~Rampa, ``Federated learning with cooperating
  devices: A consensus approach for massive iot networks,'' \emph{IEEE Internet
  of Things Journal}, vol.~7, no.~5, pp. 4641--4654, 2020.

\bibitem{cross-grad}
Y.~Esfandiari~et al., ``Cross-gradient aggregation for decentralized learning
  from non-iid data,'' in \emph{International Conference on Machine
  Learning}.\hskip 1em plus 0.5em minus 0.4em\relax PMLR, 2021, pp. 3036--3046.

\bibitem{sayed2014adaptation}
A.~H. Sayed, ``Adaptation, learning, and optimization over networks,''
  \emph{Foundations and Trends in Machine Learning}, vol.~7, pp. 311--801,
  2014.

\bibitem{plata2015distributed}
J.~Plata-Chaves, N.~Bogdanovi{\'c}, and K.~Berberidis, ``Distributed
  diffusion-based lms for node-specific adaptive parameter estimation,''
  \emph{IEEE Transactions on Signal Processing}, vol.~63, no.~13, 2015.

\bibitem{MNIST}
Y.~Lecun~et al., ``Gradient-based learning applied to document recognition,''
  \emph{Proceedings of the IEEE}, vol.~86, no.~11, pp. 2278--2324, 1998.

\bibitem{DifAdt}
J.~Chen and A.~H. Sayed, ``Diffusion adaptation strategies for distributed
  optimization and learning over networks,'' \emph{IEEE Transactions on Signal
  Processing}, vol.~60, no.~8, pp. 4289--4305, 2012.

\bibitem{SAYEDBOOKCHAPTER}
A.~H. Sayed, ``Chapter 9 - diffusion adaptation over networks,'' in
  \emph{Academic Press Library in Signal Processing: Vol. 3}.\hskip 1em plus
  0.5em minus 0.4em\relax Elsevier, 2014.

\bibitem{OLFATI-SABER}
R.~Olfati-Saber, J.~A. Fax, and R.~M. Murray, ``Consensus and cooperation in
  networked multi-agent systems,'' \emph{Proceedings of the IEEE}, vol.~95,
  no.~1, pp. 215--233, 2007.

\end{thebibliography}

\end{document}